\documentclass{article}
\usepackage{spconf,amsmath,graphicx}
\usepackage{multirow}
\usepackage{booktabs}
\usepackage[inkscapelatex=false]{svg}

\usepackage{xcolor}
\usepackage{comment}
\usepackage{hyperref}

\title{LOSS MASKING IS NOT NEEDED IN DECODER-ONLY TRANSFORMER FOR DISCRETE-TOKEN-BASED ASR}
\name{\shortstack{Qian Chen, Wen Wang, Qinglin Zhang, Siqi Zheng, Shiliang Zhang \\ \it Chong Deng, Yukun Ma, Hai Yu, Jiaqing Liu, Chong Zhang}}
\address{\{tanqing.cq,w.wang\}@alibaba-inc.com \\ Speech Lab, Alibaba Group}

\begin{document}
\ninept
\maketitle
\begin{abstract}
Recently, unified speech-text models, such as SpeechGPT, VioLA, and AudioPaLM, have achieved remarkable performance on various speech tasks. These models discretize speech signals into tokens (speech discretization) and use a shared vocabulary for both text and speech tokens. Then they train a single decoder-only Transformer on a mixture of speech tasks. However, these models rely on the Loss Masking strategy for the ASR task, which ignores the dependency among speech tokens. 
In this paper, we propose to model speech tokens in an autoregressive way, similar to text. We find that applying the conventional cross-entropy loss on input speech tokens does not consistently improve the ASR performance over the Loss Masking approach. To address this issue, we propose a novel approach denoted \textbf{Smoothed Label Distillation (SLD)}, which applies a KL divergence loss with smoothed labels on speech tokens. Our experiments show that SLD effectively models speech tokens and outperforms Loss Masking for decoder-only Transformers in ASR tasks with different speech discretization methods\footnote{The source code can be found here: \url{https://github.com/alibaba-damo-academy/SpokenNLP/tree/main/sld}}. 
\end{abstract}
\vspace{-2mm}
\section{INTRODUCTION}
\vspace{-1mm}
\label{sec:intro}
Large language models (LLMs) such as GPT-4~\cite{DBLP:journals/corr/abs-2303-08774}, PaLM2~\cite{DBLP:journals/corr/abs-2305-10403}, and LLaMA~\cite{DBLP:journals/corr/abs-2302-13971} have made great progress and achieved strong zero-shot generalization capability on various text tasks without fine-tuning. Recently, researchers have extended the ideas of text LLMs to the speech modality, proposing unified speech-text LLMs~\cite{DBLP:journals/corr/abs-2305-11000,DBLP:journals/corr/abs-2305-16107,DBLP:journals/corr/abs-2306-12925,DBLP:journals/corr/abs-2310-04673}. The key challenge for unified speech-text LLMs is that text is discrete while speech signal is continuous. 
To address this key challenge, novel methods such as SpeechGPT~\cite{DBLP:journals/corr/abs-2305-11000}, VioLA~\cite{DBLP:journals/corr/abs-2305-16107}, and AudioPaLM~\cite{DBLP:journals/corr/abs-2306-12925} use a technique known as \textit{speech discretization}, which converts continuous speech signals into discrete tokens. These discrete tokens are then merged with text tokens to form a shared vocabulary, and modeled by a unified, decoder-only Transformer architecture~\cite{Vaswani2017AttentionIA}. 
This direction of unified speech-text models based on discrete speech tokens and decoder-only Transformer has great potential in exploiting powerful text generative models, providing zero-shot generalizability and in-context learning capacity, and optimizing multi-tasking abilities in a unified framework. In this paper, we aim to improve this line of unified speech-text models based on discrete speech tokens and decoder-only Transformer.

However, current implementations of unified speech-text models such as SpeechGPT, VioLA, and AudioPaLM adopt a \textit{Loss Masking} strategy for the ASR tasks. Loss Masking overlooks the explicit modeling of dependencies between speech tokens. Despite the discretization of speech input, these models are not designed to predict speech tokens autoregressively as they do with text, which can result in underutilization of training data and an incomplete representation of dependencies among speech tokens.

To overcome this limitation, our research explores the autoregressive modeling of speech tokens. We find that the conventional cross-entropy loss on speech tokens does not improve ASR performance over Loss Masking. Consequently, we introduce a novel method called \textbf{Smoothed Label Distillation (SLD)}. This method employs a KL divergence loss with \textit{smoothed labels} to effectively model the speech tokens. Our empirical evaluations demonstrate that SLD not only alleviates the shortcomings associated with cross-entropy loss but also outperforms the Loss Masking strategy in ASR tasks with different speech discretization methods.

\vspace{-2mm}
\section{Related Work} 
\vspace{-1mm}
\label{sec:prior} 
Discrete-token-based ASR models use discrete tokens as input for ASR models~\cite{DBLP:conf/iclr/BaevskiSA20,DBLP:journals/corr/abs-1911-03912,DBLP:journals/corr/abs-2305-18108}. As summarized in~\cite{DBLP:journals/corr/abs-2305-18108}, discrete-token-based ASR has several advantages over conventional log-mel-filterbank-feature based ASR models: Discrete speech tokens can encode both acoustic and semantic information (but with less speaker-specific information), preserve original speech duration information, and reduce storage and transmission size. However, prior discrete-token-based ASR models underperform conventional ASR models using log-mel-filterbank features.

There are two main research directions on discrete-token-based ASR. One direction focuses on developing ASR models using discrete tokens and existing Transformer \textit{encoder} or \textit{encoder-decoder} architectures.
For example, in~\cite{DBLP:conf/iclr/BaevskiSA20}, a Transformer~\cite{Vaswani2017AttentionIA} is trained on vq-wav2vec Gumbel-Softmax discrete tokens. Similarly, \cite{DBLP:journals/corr/abs-1911-03912} uses a BERT encoder~\cite{DBLP:conf/naacl/DevlinCLT19} with CTC loss on the same vq-wav2vec discrete tokens. On the other hand, \cite{DBLP:journals/corr/abs-2305-18108} employs a joint CTC/attention-based encoder-decoder architecture based on the E-Branchformer~\cite{DBLP:conf/slt/KimWPPSHW22} using WavLM~\cite{DBLP:journals/jstsp/ChenWCWLCLKYXWZ22} k-means discrete tokens.
The other direction focuses on developing unified speech-text models based on discrete tokens for ASR and other speech tasks such as speech-to-text and speech-to-speech translation and text-to-speech synthesis using \textit{decode-only} Transformer architecture~\cite{DBLP:journals/corr/abs-2305-11000,DBLP:journals/corr/abs-2305-16107,DBLP:journals/corr/abs-2306-12925}. These models use discrete speech tokens to represent speech signals and employ the \textit{decoder-only} Transformer architecture. For instance, SpeechGPT \cite{DBLP:journals/corr/abs-2305-11000} utilizes LLaMA~\cite{DBLP:journals/corr/abs-2302-13971} as the backbone with HuBERT~\cite{DBLP:journals/taslp/HsuBTLSM21} k-means discrete tokens. VioLA~\cite{DBLP:journals/corr/abs-2305-16107} employs an 18-layer decoder-only Transformer using discrete codec codes based on the EnCodec model~\cite{DBLP:journals/corr/abs-2210-13438}. AudioPaLM~\cite{DBLP:journals/corr/abs-2306-12925} uses PaLM-2 \cite{DBLP:journals/corr/abs-2305-10403} as the backbone with discrete tokens extracted from the USM encoder~\cite{DBLP:journals/corr/abs-2303-01037}. Notably, all these three models use a Loss Masking strategy, which means that they do not explicitly model the dependency between the speech tokens. In this paper, we propose a new method to replace Loss Masking and effectively model the dependency between the speech tokens for models in the second direction.

\begin{figure*}[t]
\begin{minipage}[b]{.33\linewidth}
  \centering
  \centerline{\includegraphics[width=4.0cm]{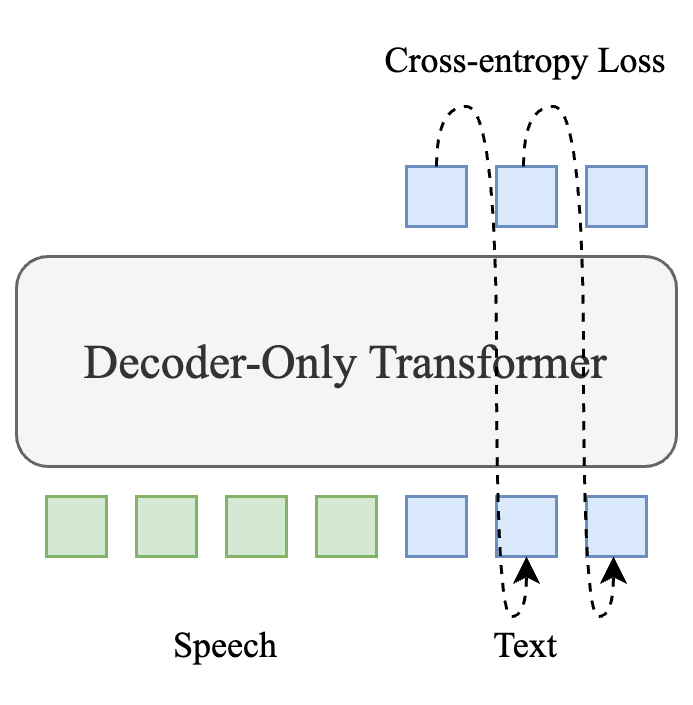}}
  \centerline{(a) Loss Masking}\medskip
\end{minipage}
\begin{minipage}[b]{.33\linewidth}
  \centering
  \centerline{\includegraphics[width=4.0cm]{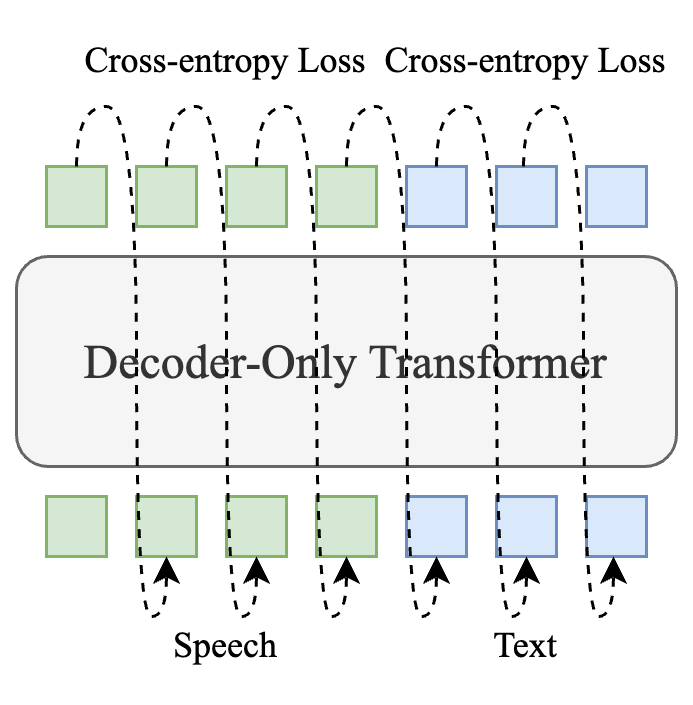}}
  \centerline{(b) Multimodal Cross-Entropy Loss}\medskip
\end{minipage}
\hfill
\begin{minipage}[b]{0.33\linewidth}
  \centering
  \centerline{\includegraphics[width=4.0cm]{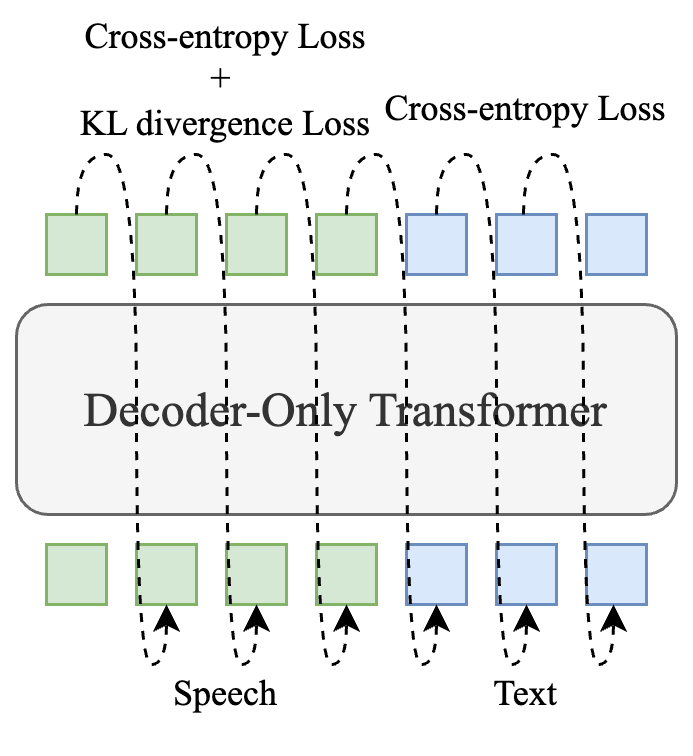}}
  \centerline{(c) Smoothed Label Distillation (SLD) (\textbf{Ours})}\medskip
\end{minipage}
\caption{The comparison of different methods for training discrete-token-based decoder-only Transformer for ASR. (1) \textbf{Loss Masking} masks the loss for speech tokens and uses cross-entropy loss for text tokens; (2) \textbf{Multimodal Cross-Entropy Loss}  uses cross-entropy loss for both speech and text tokens, based on the outputs of the model and the hard labels; (3) \textbf{Our proposed Smoothed Label Distillation (SLD)} adds a KL divergence loss between the model outputs and the \textbf{smoothed labels} for speech tokens, on top of the Multimodal Cross-Entropy Loss. }
\label{fig:main}
\end{figure*}

\section{METHOD}
\label{sec:method}

\subsection{Model Framework}
Unlike the typical encoder-decoder architecture for ASR, we utilize a decoder-only Transformer to model the ASR task, which is also employed in SpeechGPT, AudioPaLM, and VioLA. Given a speech sample $x'$ and its transcription $y'$, $x'$ is transformed into discrete tokens, for example, using the WavLM or HuBERT models with k-means, and then further tokenized using subword modeling based on unigram language model~\cite{DBLP:conf/acl/Kudo18}, resulting in $x$. The text $y'$ is tokenized using byte-pair-encoding (BPE)~\cite{DBLP:conf/acl/SennrichHB16a} to generate the sequence $y$. The ASR task aims to predict the text BPE sequence $y$ according to the discrete speech tokens $x$. The model is optimized by maximizing $p(y'|x'; \theta) = p(y|x; \theta)$, where $\theta$ denotes the model parameters.

A text-only Transformer decoder is modified to model both text and speech by changing the token embedding matrix in a Transformer decoder. The original matrix $E$ with $t$ tokens and $d$ embedding dimensions, which maps tokens to dense embeddings, is expanded to a size of $(t + s) \times d$, where $s$ is the number of speech tokens. Another embedding matrix $E'$ in the final softmax layer is also modified accordingly. By adding new rows to the pre-trained text model checkpoints, the model can be used for initializing and training new speech embeddings. All model parameters are trained, including all the weights, to ensure a high performance.

\subsection{Training Objectives}
We compare three objectives for training a discrete-token-based decoder-only Transformer for ASR, namely, Loss Masking, Multimodal cross-entropy loss, and the proposed SLD approach.

\noindent \textbf{Loss Masking} Used by SpeechGPT, VioLA, and AudioPaLM, Loss Masking does not learn the discrete speech tokens in an autoregressive manner like text. As illustrated in Fig. \ref{fig:main} (a), this approach masks the loss for speech tokens and uses cross-entropy (CE) loss only on the \textit{text BPE tokens}. 
\begin{equation}
\label{eq:text}
\mathcal{L}_{\text{CE\_text}} = - \sum_{t=1}^{T_t} \log p(y_t|x, y_{<t}; \theta)
\end{equation}
where $T_t$ is the number of the text BPE tokens, $y_t$ is the BPE token at time step $t$ and $y_{<t}$ is the BPE tokens earlier than time step $t$. The number of training samples is omitted for simplicity. As emphasized in Section~\ref{sec:intro}, Loss Masking has its drawbacks of ineffectiveness in modeling the dependency between speech tokens and wasting training data.

\noindent \textbf{Multimodal Cross-entropy Loss} 
The naive multimodal CE loss uses autoregressive CE loss for both speech tokens and text tokens based on the outputs of the model and the \textit{hard} labels, as illustrated in Fig. \ref{fig:main} (b). The potential advantage of this approach over Loss Masking is that it could exploit speech data and model dependency between speech tokens. The CE loss on discrete speech tokens is:
\begin{equation}
\mathcal{L}_{\text{CE\_speech}} = - \sum_{t=1}^{T_s} \log p(x_t|x_{<t}; \theta) 
\end{equation}
\noindent where $T_s$ is the number of discrete speech tokens, $x_t$ is the speech token at time step $t$ and $x_{<t}$ is the speech tokens earlier than time step $t$. The Multimodal CE Loss is the sum of both text cross-entropy loss and speech cross-entropy loss:
\begin{equation}
\mathcal{L}_{\text{CE\_multimodal}} = \mathcal{L}_{\text{CE\_text}} + \mathcal{L}_{\text{CE\_speech}}
\end{equation}
\noindent \textbf{Smoothed Label Distillation (SLD)} Multimodal CE Loss treats the discrete speech tokens and text tokens in the \textit{same} way. We find that compared to Loss Masking, Multimodal CE Loss does not consistently improve the ASR performance as shown in our experiments in Section~\ref{sec:experiment}. We hypothesize that the noise introduced by speech discretization (converting continuous speech signal to discrete tokens) may be a possible cause for this inconsistent performance. 
Discretization noise is an inevitable result of representing continuous speech with a finite set of discrete tokens, as discussed in~\cite{Widrow1996StatisticalTO}. It is an inherent issue in all discretization methods. Minimizing the naive cross-entropy loss on discrete speech tokens aims to maximize the log-likelihood of the target label. However, this approach can lead to overconfidence and overfitting of noisy speech labels. Thus, the ineffectiveness of the naive cross-entropy approach may stem from the noise introduced during speech discretization.
Inspired by knowledge distillation~\cite{DBLP:journals/corr/HintonVD15}, which combines the hard target loss and the soft target loss to train the student model to mimic the teacher model, we propose a novel method called Smoothed Label Distillation (SLD). We introduce a KL divergence loss with \textit{smoothed labels} on the input speech on top of the multimodal CE loss to effectively model speech tokens, as illustrated in Fig.~\ref{fig:main} (c). The KL divergence loss is computed as:
\begin{equation}
\label{eq:kl_loss}
\mathcal{L}_{\text{KL\_speech}} = \sum_{t=1}^{T_s} D_{\text{KL}}( q'(x_t|x_{<t}) || p(x_t|x_{<t}; \theta)  ) 
\end{equation}
where $D_{\text{KL}}$ is the Kullback-Leibler divergence, and $q'(x_t|x_{<t})$ is a \textbf{smoothed} label distribution inspired by the label smoothing method~\cite{DBLP:conf/cvpr/SzegedyVISW16}: 
\begin{equation}
q'(x_t|x_{<t}) = \text{softmax}((1 - \epsilon) q(x_t|x_{<t}) + \epsilon u(x_t), T)
\end{equation}
where $q(x_t|x_{<t})$ is the original distribution of labels (discrete speech tokens), $\epsilon$ is a small constant, $T$ is a temperature, and $u(x_t)$ is the uniform distribution over labels. Thus our new method SLD optimizes the following overall loss:
\begin{equation}
\label{eq:sld}
\mathcal{L}_{\text{SLD}} = \mathcal{L}_{\text{CE\_text}} + \mathcal{L}_{\text{CE\_speech}} + \alpha \mathcal{L}_{\text{KL\_speech}}
\end{equation}
where the weight $\alpha$ is a hyperparameter. 

\section{Experiments}
\label{sec:experiment}
\subsection{Experimental Setup}
\noindent \textbf{Dataset} We evaluate the proposed SLD method on the LibriSpeech corpus~\cite{DBLP:conf/icassp/PanayotovCPK15}, which is commonly used for discrete-token-based ASR systems~\cite{DBLP:journals/corr/abs-1911-03912,DBLP:journals/corr/abs-2305-18108}. We report the word error rates (WER) of different models on four sets: dev-clean, dev-other, test-clean, and test-other. The *-other sets are more challenging than the *-clean sets. We train all of our models on the full 960 hours set with speed perturbation~\cite{DBLP:conf/interspeech/KoPPK15} as data augmentation, using factor 0.9 and 1.1. We follow the settings of the state-of-the-art (SOTA) E-Branchformer model~\cite{DBLP:journals/corr/abs-2305-18108}.

\noindent \textbf{Discrete Speech Tokens} We use the same primary model as the SOTA E-Branchformer~\cite{DBLP:journals/corr/abs-2305-18108} to extract speech representations, which is WavLM-large\footnote{\url{https://github.com/microsoft/unilm/tree/master/wavlm}}. We also compare it with HuBERT-large\footnote{\url{https://dl.fbaipublicfiles.com/hubert/hubert\_large\_ll60k.pt}} for speech discretization. We apply k-means with 2000 clusters on a subset of 100 hours of data randomly sampled from the training set \textit{without data augmentation}. Then we use Sentencepiece\footnote{\url{https://github.com/google/sentencepiece}} with the unigram language model to tokenize the cluster centroids and obtain 6000 subword units. We train these subword units on the full 960-hour training set with speed perturbation data augmentation.

\noindent \textbf{Backbone} We adopt GPT-2~\cite{Radford2019LanguageMA} as the backbone of our decoder-only Transformer model for discrete-token-based ASR systems. We use the GPT2-medium model (350M parameters)\footnote{\url{https://huggingface.co/gpt2-medium}}, which has 24 layers, 1024 hidden size, 16 attention heads, 1024 max sequence length, and 50257 BPE tokens. We expand the vocabulary with the 6000 speech subword units and two special end tokens (i.e., $<$speech\_end$>$ and $<$text\_end$>$).

\noindent \textbf{Training and Inference Configuration}
We use the AdamW optimizer with a learning rate of 3e-4.
We train our models on 8 A800 80GB GPUs for 10 epochs, with batch size 64. Each setup takes approximately 1 day to finish.
We apply time masking to all input tokens, including speech and text tokens, by replacing each token with a special padding token with a probability of 0.3. We also set the dropout rate as 0.1.
We use temperature 1, the weight $\alpha$ as 0.008, and $\epsilon$ 0.1 for the KL divergence loss. For evaluation, we select the model that has the lowest WERs on a subset of dev-other\footnote{This subset has 1215 samples randomly selected from all 2939 samples. We use it to speed up the evaluation process during each epoch.}.
We also analyze the loss curves on this subset (Section~\ref{sec:analysis}).
We use greedy search for ASR decoding and do not use any additional language models in our experiments.

\subsection{Main Results}
\label{sec:main_results}
Table~\ref{tab:main} presents a comparison of the WER categorized into three groups: those from previous literature, our models using GPT-2 with WavLM discrete tokens, and those with HuBERT discrete tokens. Our GPT-2 model with WavLM discrete tokens is compared with the SOTA E-Branchformer(WavLM discrete). The results indicate that our model achieves a slightly higher WER, thus establishing it as a competitive baseline. Further analysis shows that when our GPT-2 model incorporates Loss Masking, it slightly outperforms the implementation with Multimodal CE Loss, achieving a 3\% relative WER reduction on the test-other dataset. Our proposed Smoothed Label Distillation (SLD) method further enhances performance against Loss Masking, leading to \textbf{9\%} relative WER reduction on test-clean and \textbf{4\%} WER reduction on test-other. These results outperform Multimodal CE Loss by \textbf{9\%} and \textbf{7\%} on the respective datasets. Consequently, SLD provides superior performance to the commonly used Loss Masking in models such as SpeechGPT, VioLA, and AudioPaLM, offering a solution to the issue of applying traditional CE loss to discrete speech tokens in ASR tasks. Notably, the absolute WER differences between two runs with SLD are less than 0.1 across all test and development sets, indicating that the improvements from SLD are consistent.

Designed to enhance robustness against noise in speech discretization, our SLD method is believed to be effective regardless of the discretization method employed. The second and third groups in Table~\ref{tab:main} present SLD's performance compared to Loss Masking and Multimodal CE Loss for both the HuBERT and WavLM discretization methods. The variable gains observed suggest that training with Multimodal CE Loss, which employs \textit{hard} labels, might be sensitive to noise from speech discretization, potentially causing inconsistent ASR performance. In contrast, SLD consistently outperforms Loss Masking for both HuBERT and WavLM discretizations, supporting our hypothesis and standing out as a significant enhancement over the other methods. It is noteworthy that the GPT2 model with HuBERT discrete tokens achieves superior ASR performance compared to using WavLM discrete tokens under each loss function. This finding contradicts the results from \cite{DBLP:journals/jstsp/ChenWCWLCLKYXWZ22}, which suggest that WavLM offers comparable or better ASR performance than HuBERT. This discrepancy may derive from the fact that they utilize HuBERT or WavLM as \textit{continuous features}, while in our research, HuBERT or WavLM are applied as \textit{discrete tokens} for ASR. Determining the most effective speech representation learning methods as continuous features or as discrete tokens requires further investigation.
We conduct t-tests to compare SLD with Loss Masking and Multimodal CE for WavLM and HuBERT discrete tokens. Statistically significant WER differences are found between SLD and Loss Masking for all subsets, while SLD only shows significant gains over Multimodal CE on the noisier dev/test-other subsets.  

\begin{table}[htb]
\caption{Comparison to published results in terms
of WER(\%) on Librispeech dev and test sets. \textbf{GPT2(WavLM discrete)} uses the same WavLM-large discrete tokens as the SOTA E-Branchformer. \textbf{GPT2(HuBERT discrete)} uses HuBERT-large discrete tokens. The best WERs on each set in each group are in bold.}
\label{tab:main}
\begin{center}
\scalebox{0.9}{
\begin{tabular}{ l  c  c  c c }
\toprule
\multirow{2}{*}{Method} & \multicolumn{2}{c}{dev}   & \multicolumn{2}{c}{test} \\
   & clean & other & clean & other \\ 
\midrule
Transformer (vq-wav2vec) \cite{DBLP:conf/iclr/BaevskiSA20} &  5.6 & 15.5 
 & 6.2  & 18.2 \\
BERT (vq-wav2vec) \cite{DBLP:conf/naacl/DevlinCLT19} &  4.0 & 10.9 & 4.5 & 12.1 \\
E-Branchformer & & & & \\
~~~ (FBank) \cite{DBLP:journals/corr/abs-2305-18108} & 2.5  & 6.3 & 2.6 & 6.2 \\
~~~~(WavLM continuous) \cite{DBLP:journals/corr/abs-2305-18108} & \textbf{1.9}  & \textbf{3.9} & \textbf{2.0} & \textbf{4.0} \\
~~~~(WavLM discrete) \cite{DBLP:journals/corr/abs-2305-18108}   & 2.9  & 6.8 & 3.0 & 7.0 \\
\midrule
GPT2 (WavLM discrete)  & & & &  \\
~~~~\romannumeral1. Loss Masking    & 3.2  & 7.1 & 3.3  & 7.0 \\
~~~~\romannumeral2. CE (Multimodal CE)  & 3.2  & 7.2 & 3.3 & 7.2 \\
~~~~\romannumeral3. Label Smoothing CE & 3.1  & 7.0 & 3.1  & 6.9 \\
~~~~\romannumeral4. KL Divergence & 3.0  & 6.8 & 3.1  & \textbf{6.7} \\
~~~~\romannumeral5. CE + KL Div. (\textbf{SLD}) & \textbf{2.9}  & \textbf{6.6} & \textbf{3.0}  & \textbf{6.7} \\
\midrule
GPT2 (HuBERT discrete)  & & & &  \\
~~~~\romannumeral1. Loss Masking    & 2.9  & 6.2 & 3.0  & 6.6 \\
~~~~\romannumeral2. CE (Multimodal CE)  & 2.8  & 6.1 & 2.8 & 6.4 \\
~~~~\romannumeral3. Label Smoothing CE & 2.8  & 6.3 & 3.0  & 6.6 \\
~~~~\romannumeral4. KL Divergence & 2.8 & 6.0 & 3.0  & 6.4 \\
~~~~\romannumeral5. CE + KL Div. (\textbf{SLD}) & \textbf{2.6}  & \textbf{5.8} & \textbf{2.7}  &\textbf{6.1} \\
\bottomrule
\end{tabular}
}
\end{center}
\end{table}

\subsection{Analysis}
\label{sec:analysis}
Table~\ref{tab:main} further presents ablation analysis results, focusing on evaluating the impact of different speech loss functions when applied to WavLM and HuBERT discrete speech tokens within the GPT2 framework. 
In addition to using Loss Masking (\romannumeral1), Cross-Entropy (CE) loss (\romannumeral2) in Multimodal CE, and summing Multimodal CE and KL divergence loss (\romannumeral5) in our SLD method, we also evaluate using Label Smoothing CE loss (\romannumeral3) (that is, CE loss with label smoothing~\cite{DBLP:conf/cvpr/SzegedyVISW16}) and only using KL Divergence loss (\romannumeral4) with \textit{smoothed labels} for speech loss (that is, training with $\mathcal{L}_{\text{CE\_text}} + \alpha \mathcal{L}_{\text{KL\_speech}}$ rather than Eq.~\ref{eq:sld}).
We observe that Label Smoothing CE loss (\romannumeral3) slightly outperforms Loss Masking (\romannumeral1) on WavLM but shows comparable performance on HuBERT. 
On the other hand, KL Divergence loss (\romannumeral4) improves performance over Label Smoothing CE loss (\romannumeral3) and Loss Masking (\romannumeral1) for both WavLM and HuBERT; KL Divergence loss (\romannumeral4) outperforms Multimodel CE (\romannumeral2) for WavLM and performs comparably to Multimodel CE on HuBERT. Our proposed SLD method achieves the best ASR performance, indicating that the \textit{KL divergence loss} with \textit{smoothed labels} effectively mitigates the limitations of conventional cross-entropy loss. The \textit{combination of Multimodal CE loss and KL divergence loss} in SLD allows for \textit{label matching on text} and \textit{logit matching on speech}, which may be more effective than using label matching for both speech and text and lead to improved performance and generalization compared to using either loss alone.

\begin{figure}[t]
\centerline{\includegraphics[width=8.0cm]{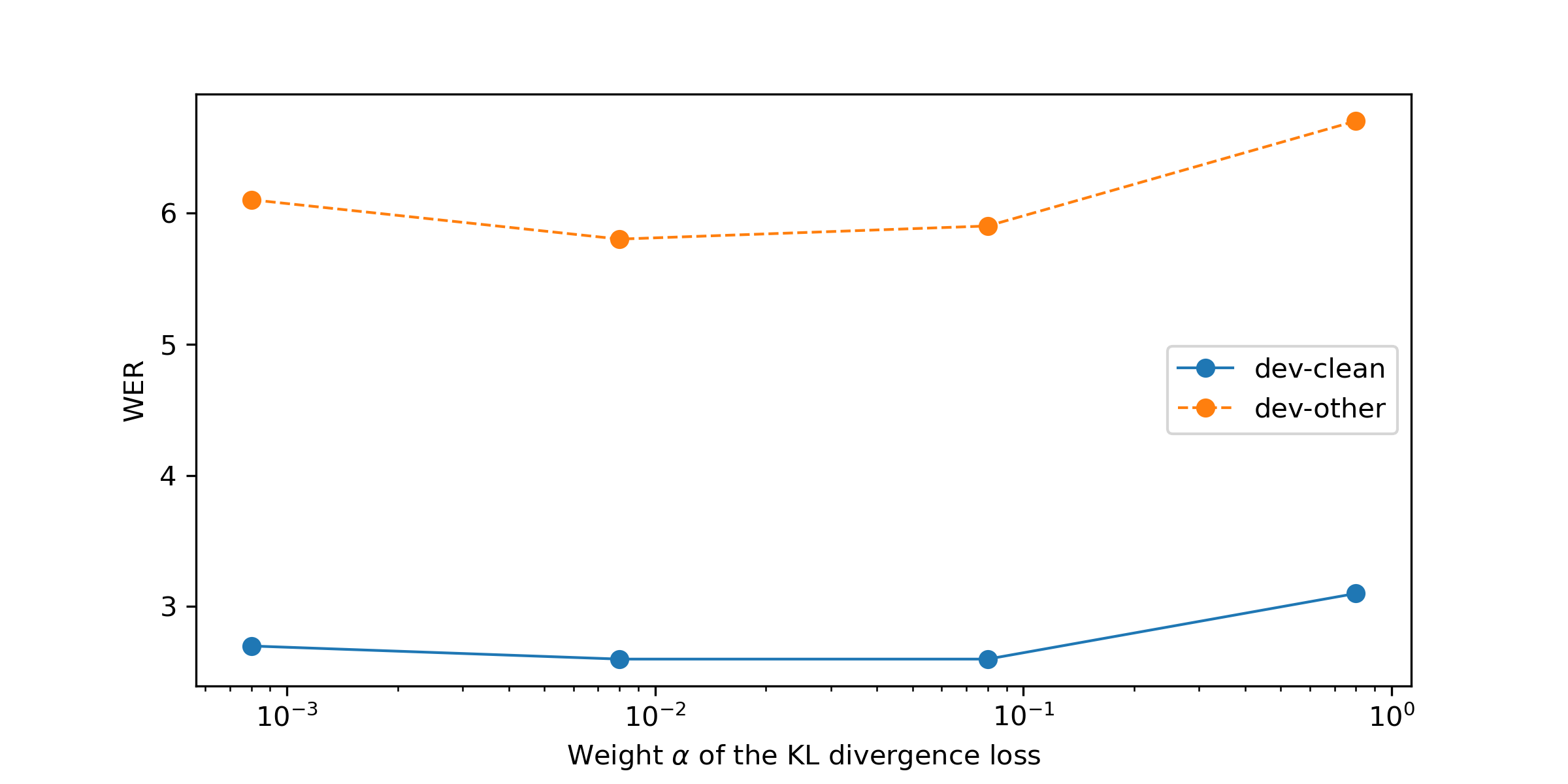 }}
\caption{The relationship between the WER of the dev-clean and dev-other datasets and the varying weight parameter $\alpha$ for the KL divergence loss component.}
\label{fig:alpha}
\end{figure}

Fig.~\ref{fig:alpha} illustrates the correlation between WER for the dev-clean and dev-other datasets and the weight parameter $\alpha$ in our SLD method on HuBERT discrete tokens. Data points are represented by dotted markers, each denoting one of four specific $\alpha$ values: 0.0008, 0.008, 0.08, and 0.8. We observe that among $\alpha$ values of 0.0008, 0.008, and 0.08, our SLD outperforms Loss Masking, except for the 0.8 setting. Through our analysis, we find that $\alpha$ as 0.008 yields the best results and set it as the default for $\alpha$.

\begin{figure}[t]
\centerline{\includegraphics[width=8.0cm]{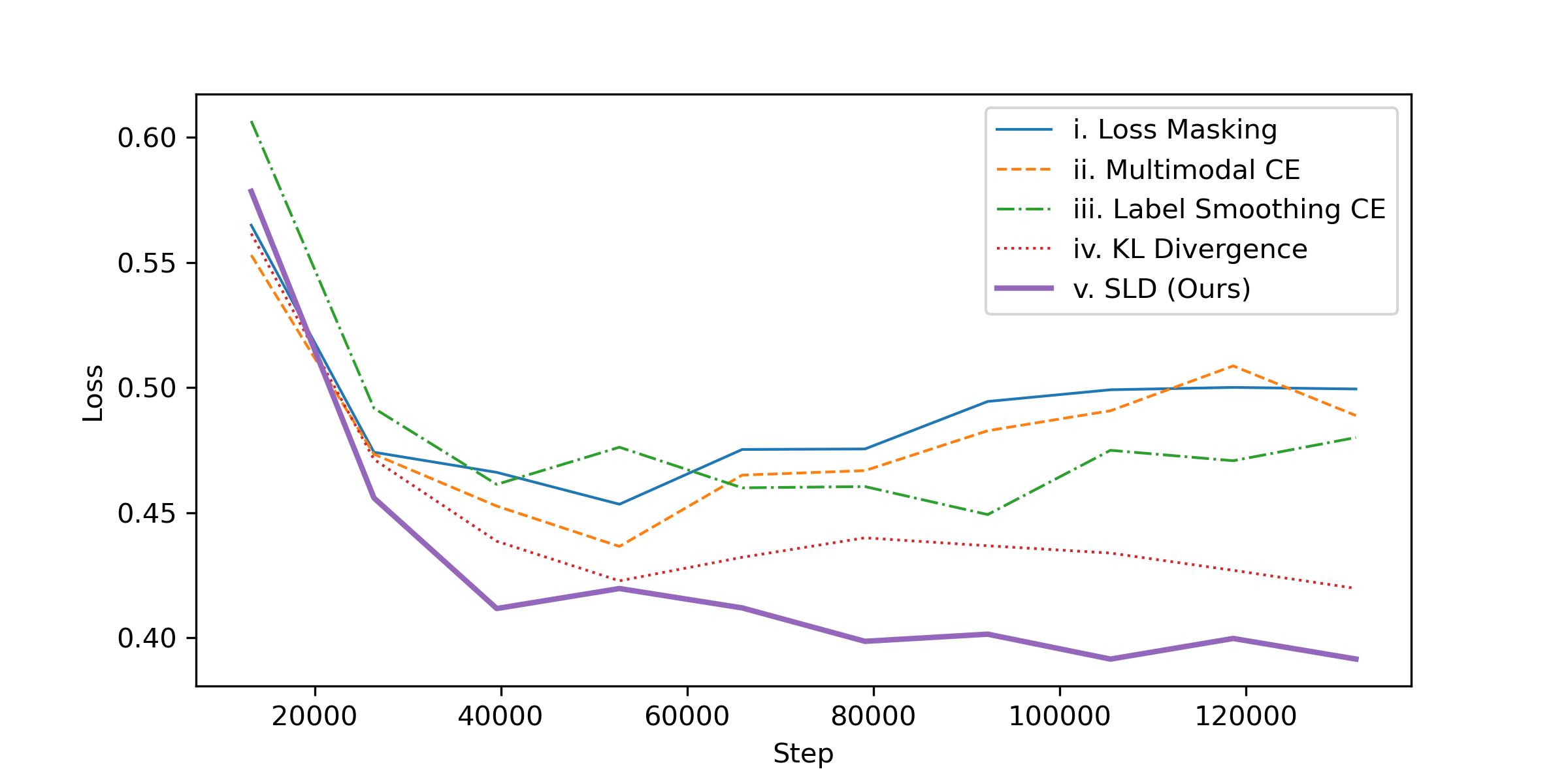}}
\caption{Comparison of loss curves for $\mathcal{L}_{\text{CE\_text}}$ on dev-other.}
\label{fig:curve}
\end{figure}

Fig.~\ref{fig:curve} illustrates loss curves for $\mathcal{L}_{\text{CE\_text}}$ on dev-other using HuBERT discrete tokens. The curves correspond to methods in Table~\ref{tab:main}, including Loss Masking, Multimodal CE, Label Smoothing CE, KL Divergence, and our SLD approach. We observe that SLD achieves the lowest final loss, followed by KL Divergence with the second lowest final loss. On the other hand, Loss Masking, Multimodal CE, and Label Smoothing CE exhibit similar higher final losses. This trend aligns with the findings in Table~\ref{tab:main}, where SLD demonstrates better performance in predicting ASR text under the teacher-forcing mode, resulting in improved ASR performance during greedy search decoding. 
We hypothesize that the lower loss curve from SLD may be attributed to its effective modeling of dependency between speech tokens and more efficient use of training data.

\section{Conclusion}
\label{sec:conclusion}
We investigate different speech losses for unified speech-text models for ASR. We find that the conventional CE loss on input speech cannot consistently improve the ASR performance over Loss Masking commonly used in existing unified speech-text ASR models. We propose a novel approach that combines the CE loss with a KL divergence loss with smoothed labels on the input speech. Experimental results on LibriSpeech demonstrate that our approach consistently outperforms loss masking on different discretization methods. We believe our approach is applicable to other speech understanding and generation tasks using unified speech-text models and will investigate it in future work.



\small
\bibliographystyle{IEEEbib}
\bibliography{refs}

\begin{thebibliography}{10}

\bibitem{DBLP:journals/corr/abs-2303-08774}
OpenAI,
\newblock ``{GPT-4} technical report,''
\newblock {\em CoRR}, vol. abs/2303.08774, 2023.

\bibitem{DBLP:journals/corr/abs-2305-10403}
Rohan Anil, Andrew~M. Dai, Orhan Firat, and et~al.,
\newblock ``Palm 2 technical report,''
\newblock {\em CoRR}, vol. abs/2305.10403, 2023.

\bibitem{DBLP:journals/corr/abs-2302-13971}
Hugo Touvron, Thibaut Lavril, Gautier Izacard, Xavier Martinet, Marie{-}Anne Lachaux, Timoth{\'{e}}e Lacroix, Baptiste Rozi{\`{e}}re, Naman Goyal, Eric Hambro, Faisal Azhar, Aur{\'{e}}lien Rodriguez, Armand Joulin, Edouard Grave, and Guillaume Lample,
\newblock ``Llama: Open and efficient foundation language models,''
\newblock {\em CoRR}, vol. abs/2302.13971, 2023.

\bibitem{DBLP:journals/corr/abs-2305-11000}
Dong Zhang, Shimin Li, Xin Zhang, Jun Zhan, Pengyu Wang, Yaqian Zhou, and Xipeng Qiu,
\newblock ``Speechgpt: Empowering large language models with intrinsic cross-modal conversational abilities,''
\newblock {\em CoRR}, vol. abs/2305.11000, 2023.

\bibitem{DBLP:journals/corr/abs-2305-16107}
Tianrui Wang, Long Zhou, Ziqiang Zhang, Yu~Wu, Shujie Liu, Yashesh Gaur, Zhuo Chen, Jinyu Li, and Furu Wei,
\newblock ``Viola: Unified codec language models for speech recognition, synthesis, and translation,''
\newblock {\em CoRR}, vol. abs/2305.16107, 2023.

\bibitem{DBLP:journals/corr/abs-2306-12925}
Paul~K. Rubenstein, Chulayuth Asawaroengchai, Duc~Dung Nguyen, and et~al.,
\newblock ``Audiopalm: {A} large language model that can speak and listen,''
\newblock {\em CoRR}, vol. abs/2306.12925, 2023.

\bibitem{DBLP:journals/corr/abs-2310-04673}
Jiaming Wang, Zhihao Du, Qian Chen, Yunfei Chu, Zhifu Gao, Zerui Li, Kai Hu, Xiaohuan Zhou, Jin Xu, Ziyang Ma, Wen Wang, Siqi Zheng, Chang Zhou, Zhijie Yan, and Shiliang Zhang,
\newblock ``Lauragpt: Listen, attend, understand, and regenerate audio with {GPT},''
\newblock {\em CoRR}, vol. abs/2310.04673, 2023.

\bibitem{Vaswani2017AttentionIA}
Ashish Vaswani, Noam~M. Shazeer, Niki Parmar, Jakob Uszkoreit, Llion Jones, Aidan~N. Gomez, Lukasz Kaiser, and Illia Polosukhin,
\newblock ``Attention is all you need,''
\newblock in {\em NIPS}, 2017.

\bibitem{DBLP:conf/iclr/BaevskiSA20}
Alexei Baevski, Steffen Schneider, and Michael Auli,
\newblock ``vq-wav2vec: Self-supervised learning of discrete speech representations,''
\newblock in {\em 8th International Conference on Learning Representations, {ICLR} 2020, Addis Ababa, Ethiopia, April 26-30, 2020}. 2020, OpenReview.net.

\bibitem{DBLP:journals/corr/abs-1911-03912}
Alexei Baevski, Michael Auli, and Abdelrahman Mohamed,
\newblock ``Effectiveness of self-supervised pre-training for speech recognition,''
\newblock {\em CoRR}, vol. abs/1911.03912, 2019.

\bibitem{DBLP:journals/corr/abs-2305-18108}
Xuankai Chang, Brian Yan, Yuya Fujita, Takashi Maekaku, and Shinji Watanabe,
\newblock ``Exploration of efficient end-to-end {ASR} using discretized input from self-supervised learning,''
\newblock {\em CoRR}, vol. abs/2305.18108, 2023.

\bibitem{DBLP:conf/naacl/DevlinCLT19}
Jacob Devlin, Ming{-}Wei Chang, Kenton Lee, and Kristina Toutanova,
\newblock ``{BERT:} pre-training of deep bidirectional transformers for language understanding,''
\newblock in {\em Proceedings of the 2019 Conference of the North American Chapter of the Association for Computational Linguistics: Human Language Technologies, {NAACL-HLT} 2019, Minneapolis, MN, USA, June 2-7, 2019, Volume 1 (Long and Short Papers)}, Jill Burstein, Christy Doran, and Thamar Solorio, Eds. 2019, pp. 4171--4186, Association for Computational Linguistics.

\bibitem{DBLP:conf/slt/KimWPPSHW22}
Kwangyoun Kim, Felix Wu, Yifan Peng, Jing Pan, Prashant Sridhar, Kyu~Jeong Han, and Shinji Watanabe,
\newblock ``E-branchformer: Branchformer with enhanced merging for speech recognition,''
\newblock in {\em {IEEE} Spoken Language Technology Workshop, {SLT} 2022, Doha, Qatar, January 9-12, 2023}. 2022, pp. 84--91, {IEEE}.

\bibitem{DBLP:journals/jstsp/ChenWCWLCLKYXWZ22}
Sanyuan Chen, Chengyi Wang, Zhengyang Chen, Yu~Wu, Shujie Liu, Zhuo Chen, Jinyu Li, Naoyuki Kanda, Takuya Yoshioka, Xiong Xiao, Jian Wu, Long Zhou, Shuo Ren, Yanmin Qian, Yao Qian, Jian Wu, Michael Zeng, Xiangzhan Yu, and Furu Wei,
\newblock ``Wavlm: Large-scale self-supervised pre-training for full stack speech processing,''
\newblock {\em {IEEE} J. Sel. Top. Signal Process.}, vol. 16, no. 6, pp. 1505--1518, 2022.

\bibitem{DBLP:journals/taslp/HsuBTLSM21}
Wei{-}Ning Hsu, Benjamin Bolte, Yao{-}Hung~Hubert Tsai, Kushal Lakhotia, Ruslan Salakhutdinov, and Abdelrahman Mohamed,
\newblock ``Hubert: Self-supervised speech representation learning by masked prediction of hidden units,''
\newblock {\em {IEEE} {ACM} Trans. Audio Speech Lang. Process.}, vol. 29, pp. 3451--3460, 2021.

\bibitem{DBLP:journals/corr/abs-2210-13438}
Alexandre D{\'{e}}fossez, Jade Copet, Gabriel Synnaeve, and Yossi Adi,
\newblock ``High fidelity neural audio compression,''
\newblock {\em CoRR}, vol. abs/2210.13438, 2022.

\bibitem{DBLP:journals/corr/abs-2303-01037}
Yu~Zhang, Wei Han, James Qin, and et~al.,
\newblock ``Google {USM:} scaling automatic speech recognition beyond 100 languages,''
\newblock {\em CoRR}, vol. abs/2303.01037, 2023.

\bibitem{DBLP:conf/acl/Kudo18}
Taku Kudo,
\newblock ``Subword regularization: Improving neural network translation models with multiple subword candidates,''
\newblock in {\em Proceedings of the 56th Annual Meeting of the Association for Computational Linguistics, {ACL} 2018, Melbourne, Australia, July 15-20, 2018, Volume 1: Long Papers}, Iryna Gurevych and Yusuke Miyao, Eds. 2018, pp. 66--75, Association for Computational Linguistics.

\bibitem{DBLP:conf/acl/SennrichHB16a}
Rico Sennrich, Barry Haddow, and Alexandra Birch,
\newblock ``Neural machine translation of rare words with subword units,''
\newblock in {\em Proceedings of the 54th Annual Meeting of the Association for Computational Linguistics, {ACL} 2016, August 7-12, 2016, Berlin, Germany, Volume 1: Long Papers}. 2016, The Association for Computer Linguistics.

\bibitem{Widrow1996StatisticalTO}
Bernard Widrow, Istv{\'a}n Koll{\'a}r, and Ming-Chang Liu,
\newblock ``Statistical theory of quantization,''
\newblock {\em IEEE Transactions on Instrumentation and Measurement}, vol. 45, pp. 353--361, 1996.

\bibitem{DBLP:journals/corr/HintonVD15}
Geoffrey~E. Hinton, Oriol Vinyals, and Jeffrey Dean,
\newblock ``Distilling the knowledge in a neural network,''
\newblock {\em CoRR}, vol. abs/1503.02531, 2015.

\bibitem{DBLP:conf/cvpr/SzegedyVISW16}
Christian Szegedy, Vincent Vanhoucke, Sergey Ioffe, Jonathon Shlens, and Zbigniew Wojna,
\newblock ``Rethinking the inception architecture for computer vision,''
\newblock in {\em 2016 {IEEE} Conference on Computer Vision and Pattern Recognition, {CVPR} 2016, Las Vegas, NV, USA, June 27-30, 2016}. 2016, pp. 2818--2826, {IEEE} Computer Society.

\bibitem{DBLP:conf/icassp/PanayotovCPK15}
Vassil Panayotov, Guoguo Chen, Daniel Povey, and Sanjeev Khudanpur,
\newblock ``Librispeech: An {ASR} corpus based on public domain audio books,''
\newblock in {\em 2015 {IEEE} International Conference on Acoustics, Speech and Signal Processing, {ICASSP} 2015, South Brisbane, Queensland, Australia, April 19-24, 2015}. 2015, pp. 5206--5210, {IEEE}.

\bibitem{DBLP:conf/interspeech/KoPPK15}
Tom Ko, Vijayaditya Peddinti, Daniel Povey, and Sanjeev Khudanpur,
\newblock ``Audio augmentation for speech recognition,''
\newblock in {\em {INTERSPEECH} 2015, 16th Annual Conference of the International Speech Communication Association, Dresden, Germany, September 6-10, 2015}. 2015, pp. 3586--3589, {ISCA}.

\bibitem{Radford2019LanguageMA}
Alec Radford, Jeff Wu, Rewon Child, David Luan, Dario Amodei, and Ilya Sutskever,
\newblock ``Language models are unsupervised multitask learners,''
\newblock in {\em OpenAI blog}, 2019.

\end{thebibliography}

\end{document}